\documentclass[twoside]{article}

\usepackage[accepted]{aistats2023}
\usepackage{amsmath}
\usepackage{amssymb}
\usepackage{graphicx}
\usepackage{hyperref}
\usepackage{mathtools}
\usepackage[round]{natbib}

\begin{document}



\twocolumn[

\aistatstitle{Hierarchical-Hyperplane Kernels for Actively Learning Gaussian Process Models of Nonstationary Systems}

\aistatsauthor{ Matthias Bitzer \And Mona Meister \And  Christoph Zimmer }

\aistatsaddress{Bosch Center for Artificial\\ Intelligence, Renningen, Germany \And  Bosch Center for Artificial\\ Intelligence, Renningen, Germany \And Bosch Center for Artificial\\ Intelligence, Renningen, Germany} ]

\begin{abstract}
	Learning precise surrogate models of complex computer simulations and physical machines often require long-lasting or expensive experiments. Furthermore, the modeled physical dependencies exhibit nonlinear and nonstationary behavior. Machine learning methods that are used to produce the surrogate model should therefore address these problems by providing a scheme to keep the number of queries small, e.g. by using active learning and be able to capture the nonlinear and nonstationary properties of the system. One way of modeling the nonstationarity is to induce input-partitioning, a principle that has proven to be advantageous in active learning for Gaussian processes. However, these methods either assume a known partitioning, need to introduce complex sampling schemes or rely on very simple geometries. In this work, we present a simple, yet powerful kernel family that incorporates a partitioning that: i) is learnable via gradient-based methods, ii) uses a geometry that is more flexible than previous ones, while still being applicable in the low data regime. Thus, it provides a good prior for active learning procedures. We empirically demonstrate excellent performance on various active learning tasks. 
\end{abstract}

\section{INTRODUCTION}
Active learning is a principled way to learn a model in a sequential data-efficient manner. It is especially useful when the collection of data is expensive. For classification, this is the case for the manual labeling procedure \citep{settles2009active}. Regression tasks in which active learning is used are, e.g. the learning of a surrogate model of complex physical processes like complex machines \citep{SafeALTimeSeries} or surrogate modeling of long-running computer simulations \citep{gramacy2020surrogates}. Thereby, queries to the oracle are either very expensive/energy intensive or take a very long time, which makes it necessary to minimize the number of queries. Recent studies on active learning for regression tasks utilized Gaussian processes \citep{HennigGarnett,SchreiterSafeAL,WaMIGp,shapeControl,li2022safe}, which have the great advantage of providing a principled notion of uncertainty, making them ideal candidates for active learning algorithms.

In practice, Gaussian process regression is often used with stationary kernels, such as the Squared Exponential Kernel. However, the stationarity of the kernel implicitly assumes that the correlation of the function values of the learned function is translation invariant. This assumption is often not met in practice and major performance gains were shown for passive learning using nonstationary Gaussian processes \citep{NIPS2017_c65d7bd7,treedGP,DKL}. Furthermore, when active learning is employed, the stationarity assumption implies an almost uniform input design [see \cite{WaMIGp}]. Nonstationary kernels on the other hand induce a non-uniform sampling in the input space, which has been shown to be beneficial when the data actually exhibits nonstationarities \citep{gpTreedAL,WaMIGp}.

In particular, input-partitioning showed promising results when combined with active learning \citep{PartitionedAL,gpTreedAL,NonMyopicAL}. However, existing methods either rely on a fixed, known partition \citep{PartitionedAL,NonMyopicAL} or use restricted geometries/ priors in function space combined with complex sampling schemes \citep{gpTreedAL}. Our goal is to provide a simple, yet powerful partitioning kernel that can be used as plug-and-play in most Gaussian process frameworks with the same (or even better) sampling behavior in active learning procedures. Concretely, our partitioning is constructed via a hierarchy of hyperplanes, build with sigmoidal gates to introduce differentiability of the kernel and smoothness of the function prior. Furthermore, the geometry of the partitioning is more flexible compared to the input-aligned partitions of \cite{gpTreedAL}, but still simple enough for the usage in the low data regime. In the next section, we give an overview over related work. In Section 3 we introduce our method/kernel and give some motivation for its usage in active learning settings. Finally, in the experimental section, we compare against different nonstationary Gaussian process based models on several real-world active learning tasks and show excellent performance.

\section{BACKGROUND AND RELATED WORK}

\paragraph{Gaussian Processes.} 

Gaussian processes provide expressive priors over functions that can be used for surrogate modeling and regression tasks. Formally, for $\mathcal{X}\subset \mathbb{R}^{d}$, a Gaussian process (GP) is a probability distribution over functions $f:\mathcal{X}\to \mathbb{R}$ for which each finite selection of function values $[f(x_{t})]_{t=1}^{T}$ at input points $x_{1},\dots,x_{T}$ has a multivariate normal distribution. The GP is fully characterized by its mean function $\mu:\mathcal{X}\to \mathbb{R}$ and its covariance function $k:\mathcal{X}\times\mathcal{X} \to \mathbb{R}_{+}$ also called the kernel. The kernel incorporates the major properties of the resulting sample functions and is often parameterized with some parameters $\Theta$. For regression, a dataset $\mathcal{D}:=\{\mathbf{x}_{T},\mathbf{y}_{T}\}$ with inputs $\mathbf{x}_{T}=[x_{1},\dots,x_{T}]$ and ouputs $\mathbf{y}_{T}=[y_{1},\dots,y_{T}]$ is considered, where the observations are perturbed with Gaussian noise $y_{t}=f(x_{t})+\epsilon_{t}$ with $\epsilon_{t}\sim \mathcal{N}(0,\sigma^{2})$. A major advantage of the Gaussian process is that the posterior $f|\mathcal{D}$ is again a Gaussian process with closed-form expressions for the mean and kernel function
\begin{equation*}
\begin{aligned}
\label{eq:post_gp}
\mu_{T}(x)&=\mu(x)+\mathbf{k}_{T}(x)^{\intercal}(K_{T}+\sigma^{2}I)^{-1}(\mathbf{y}_{T}-\mu(\mathbf{x}_{T})),\\
k_{T}(x,y)&=k(x,y)-\mathbf{k}_{T}(x)^{\intercal}(K_{T}+\sigma^{2}I)^{-1}\mathbf{k}_{T}(y),
\end{aligned}
\end{equation*}
where $\mathbf{k}_{T}(x)=[k(x,x_{1}),\dots,k(x,x_{T})]^{\intercal}$ and $K_{T}=[k(x_{t},x_{l})]_{t,l=1}^{T}$. Thus, the predictive distribution for a new point $x^{*}\in\mathcal{X}$ can also be written in closed form by
\begin{equation*}
\label{eq:pred_dist}
p(f^{*}|x^{*},\mathcal{D},\Theta)=\mathcal{N}(\mu_{T}(x^{*}),\sigma_{T}^{2}(x^{*})),
\end{equation*}
where $f^{*}:=f(x^{*})$ and $\sigma_{T}^{2}(x^{*}):=k_{T}(x^{*},x^{*})$. We denote the dependence on $\Theta$ if necessary. For more details on GP regression, we refer the interested reader to \citet{3569}.

\paragraph{Nonstationary GP's and Input-Partitioning.} 

In GP regression, the learned function $f$ is assumed to be a sample from a GP with kernel $k_{\Theta}$. Herein, the kernel provides the main a priori assumption on the learned function $f$. The most popular kernels such as the RBF and the Mat\'{e}rn kernel are \textit{stationary}, that means $k_{\Theta}(x,y)=k_{\Theta}(x-y)$ for all $x,y\in\mathcal{X}$. The correlation between function values therefore is translation invariant, and the modeled function is assumed to behave similarly over the complete input region. Different kinds of nonstationary kernels have been proposed so far. For time-warped kernels \citep{WaMIGp,DKL,inputWarpedBO} the input is transformed with a nonlinear mapping and chained with a stationary kernel. Further methods render parameters of stationary kernels input-dependent like input-dependent kernel variance and input-dependent lengthscale \citep{pmlr-v51-heinonen16,NIPS2017_c65d7bd7,pmlr-v51-herlands16}. For many proposed kernels [as in \citet{pmlr-v51-heinonen16} or \citet{pmlr-v51-herlands16}] the goal is to provide flexible priors and scalable inference, with priors not explicitly dedicated to the low-data regime. A further principle to induce nonstationarity is the partitioning of the input space, as for example done in \citet{treedGP}, who provide a model named TreedGP or in \cite{NonMyopicAL} and \cite{AutomaticStatistician} who provide input-partitioning on one-dimensional datasets via change-points. We present a model that can be viewed as a multi-dimensional generalization of change-points, where we replace change-points with change-hyperplanes. The resulting inductive bias in $d$ dimensions of our model is most similar to the TreedGP model. Technically, TreedGP is a Bayesian CART (Classification and Regression Tree) model with independent GP's in its leaves. However, rather than using a CART model, we employ input-dependent weighting \citep{pmlr-v51-herlands16,NonMyopicAL} to induce a non-axis aligned partitioning which at the same time has the advantage that the kernel is differentiable with respect to all its parameters and that the associated GP induces continuous sample paths.

\paragraph{Active Learning with Gaussian Processes.}

In active learning, data is selected sequentially, often guided by the current model state, which can drastically reduce the number of evaluations. We focus on active learning for regression tasks which has applications in industry, e.g. surrogate modeling of combustion engines \citep{SafeALTimeSeries}, shape control \citep{shapeControl} and in the approximation of long-running computer simulations \citep{gramacy2020surrogates}. Gaussian processes are a natural choice to pursue active learning for regression due to their principled uncertainty quantification [see \cite{SafeALTimeSeries,gramacy2020surrogates,shapeControl,Guestrin,WaMIGp}]. In case the underlying ground-truth system can be better described via a nonstationary Gaussian process, using a nonstationary kernel also has profound impact on the sample selection as the samples are not selected homogenous over the input space [see \cite{NonMyopicAL,WaMIGp,gpTreedAL}]. In our experimental section, we will compare against other nonstationary GP priors that were proposed to be used in active-learning settings, such as the TreedGP model \citep{gpTreedAL} or an input-warped GP \citep{WaMIGp}. Furthermore, \citet{sauer2020active} investigated the application of DeepGPs in active-learning settings. We will stick to the natural way of doing active-learning via querying the point with the highest information gain between the observation and all uncertain variables in the model [see \cite{ALM,BALD}] to investigate which impact our prior has on the active-learning performance.  In Section \hyperref[section::experiments]{4}, we show superior performance in terms of RMSE curves compared to the main competitors on various tasks.

\section{METHOD}
\begin{figure}
	%\vspace{.3in}
	\centering
	\includegraphics[scale=0.35]{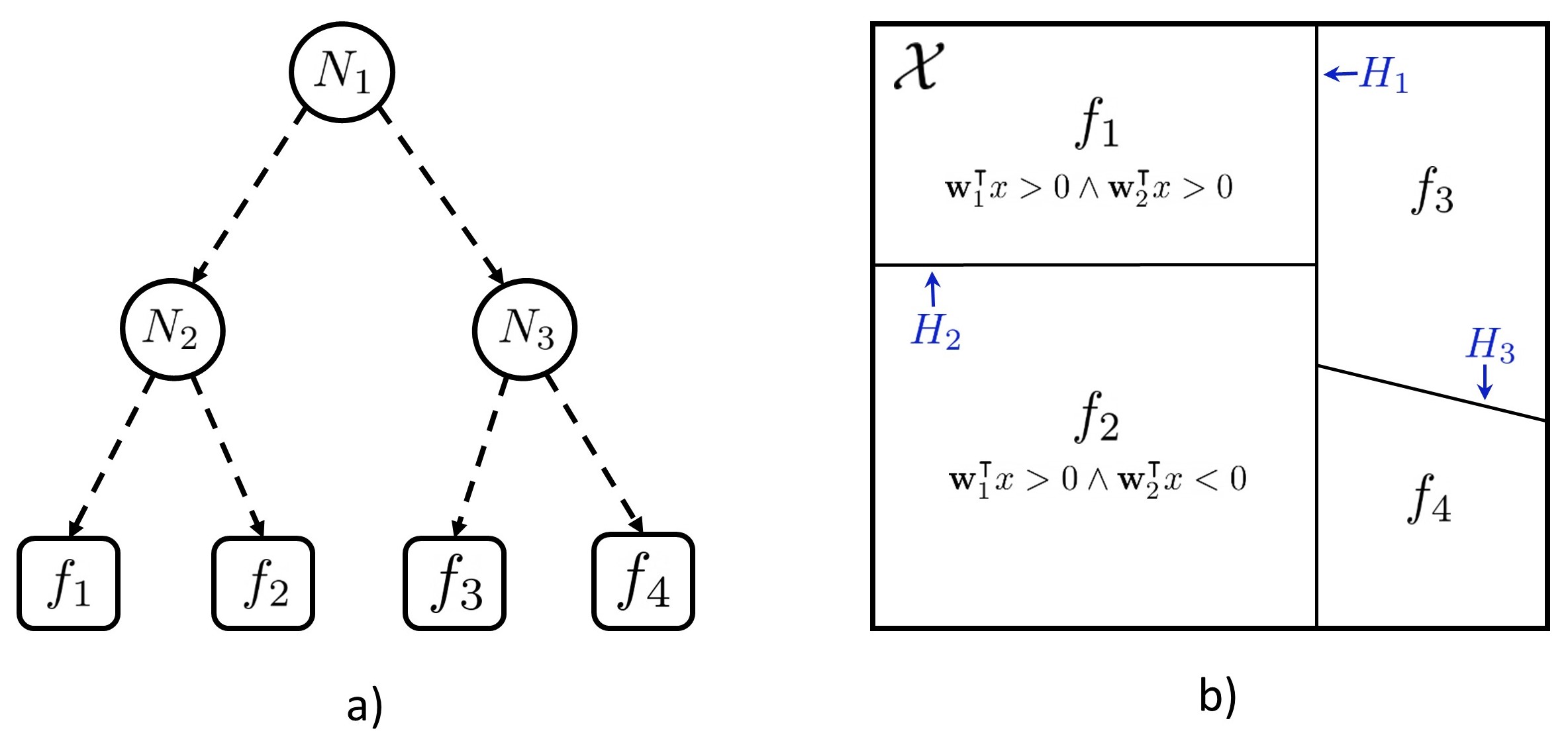}
	\vspace{.1in}
	\caption{Partitioning: a) Binary (symmetric) tree for partitioning with four leaves. b) Possible partition of the input space induced by that tree.}
	\label{fig:forpapertree}
\end{figure}
We induce input-partitioning by defining the final GP $f$ on $\mathcal{X}\subset \mathbb{R}^{d}$ as an input-dependent sum of $J$ independent latent GPs $f_{1},\dots,f_{J}$, i.e.
$$
f(x)=\sum_{j=1}^{J}\lambda_{j}(x)f_{j}(x),
$$
where $\lambda_{j}\colon \mathcal{X}\to [0,1]$ are weighting functions with $\sum_{j=1}^{J}\lambda_{j}(x)=1$ for all $x\in\mathcal{X}$. The latent GPs are  equipped with stationary kernels $k_{j}(x,y)$. The function $f$ is therefore a GP itself with kernel 
$$
k(x,y)=\sum_{j=1}^{J}\lambda_{j}(x)\lambda_{j}(y)k_{j}(x,y).
$$
We present a new kernel of this form that partitions the input hierarchically through $\lambda_{j}$.
The weighting functions $\lambda_{j}, j=1,\dots,J$, specify the regions in which the modeled function $f$ is described by the corresponding latent functions $f_{j}$. If, for example, the weighting function $\lambda_{j}(x)$ is close to one for all $x$ in some region $A\subset \mathcal{X}$, then the GP $f$ is described in that region by the kernel $k_{j}(x,y)$. Therefore, the GP behaves like a stationary GP in region $A$. 
The proposed geometry of the partition via a hierarchy of hyperplanes is inspired by a mixture of linear experts model \citep{MixtureOfLinearExperts}. In this work, we use the partitioning logic to define a nonstationary GP.

\subsection{Input-Partitioning}

The partitioning is done along a binary tree $\mathcal{T}$ with $M\coloneqq J-1$ nodes and $J$ leaves. Each node $N_{i}, i=1,\dots,M$, is associated with a vector $\mathbf{w}_{i}\in\mathbb{R}^{d+1}$ and an induced hyperplane $H_{i}=\{x\in \mathbb{R}^{d} | \mathbf{w}_{i}^{\intercal}\tilde{x}=0\}$ where $\tilde{x}=(1,x)^{\intercal}\in\mathbb{R}^{d+1}$. Each leaf represents one latent GP $f_{j}$ [see Figure \ref{fig:forpapertree}a) for an example]. Every node $N_{i}$ splits the input region with its hyperplane by placing an input-dependent weight $\sigma(\mathbf{w}_{i}^{\intercal}\tilde{x})$ to its left subtree and all associated GPs and $1-\sigma(\mathbf{w}_{i}^{\intercal}\tilde{x})$ to its right subtree, where $\sigma(\cdot)$ is the standard logistic sigmoid function. The final weights $\lambda_{j}(x), j=1,\dots,J,$ are given by multiplying all weights along their respective path in the tree:
\begin{align}
\label{weighting_func}
\lambda_{j}(x)=\prod_{i=1}^{M}\sigma(\mathbf{w}_{i}^{\intercal}\tilde{x})^{\xi_{L}(j,i)}(1-\sigma(\mathbf{w}_{i}^{\intercal}\tilde{x}))^{\xi_{R}(j,i)},
\end{align}
where $\xi_{L},\xi_{R}\colon\{1,\dots,J\}\times\{1,\dots,M\} \to \{0,1\}$ 
%and $\xi_{R}\colon\{1,\dots,J\}\times\{1,\dots,M\} \to \{0,1\}$ 
encode the tree structure with
\begin{align*}
\xi_{L}(j,i)=\begin{cases}
1~~~\text{if}~f_{j}\text{ is in the left subtree of } N_{i},\\
0~~~\text{else},\\
\end{cases}
\end{align*}
and
\begin{align*}
\xi_{R}(j,i)=\begin{cases}
1~~~\text{if}~f_{j}\text{ is in the right subtree of } N_{i},\\
0~~~\text{else}.\\
\end{cases}
\end{align*}
This weighting function leads to a hierarchical partitioning of the input space. This can be understood by observing that the multiplication of the sigmoid functions acts as a soft version of logicals AND's, where the weight $\lambda_{j}(x)$ is large whenever the input $x \in \mathcal{X}$ lies at the correct side of each hyperplane when traversing down the tree to $f_{j}$. Here, correct means either on one or the other side of the hyperplane, depending if the path in the tree progresses on the right or the left subtree. We call the resulting kernel Hierarchical-Hyperplane Kernel (HHK).

An illustrative example of the partitioning is given in Figure \ref{fig:forpapertree} a) and b). The weighting functions recursively divide the input space $\mathcal{X}=:A_{1}$ beginning with the hyperplane $H_{1}$ at node $N_{1}$ into sets $A_{2}\coloneqq\{x\in A_{1}|\mathbf{w}_{1}^{\intercal}\tilde{x}>0\}$ and $A_{3}\coloneqq\{x\in A_{1}|\mathbf{w}_{1}^{\intercal}\tilde{x}<0\}$. In the next layer, at node $N_{i}$, the associated set $A_{i}$ is again divided into the sets $\{x\in A_{i}|\mathbf{w}_{i}^{\intercal}\tilde{x}>0\}$ and $\{x\in A_{i}|\mathbf{w}_{i}^{\intercal}\tilde{x}<0\}$. The actual weighting of the stationary kernels is soft due to the application of the sigmoid functions. The partitioning can be rendered finer by increasing the tree size.

\begin{figure*}[t]
	\centering
	\includegraphics[width=0.93\linewidth]{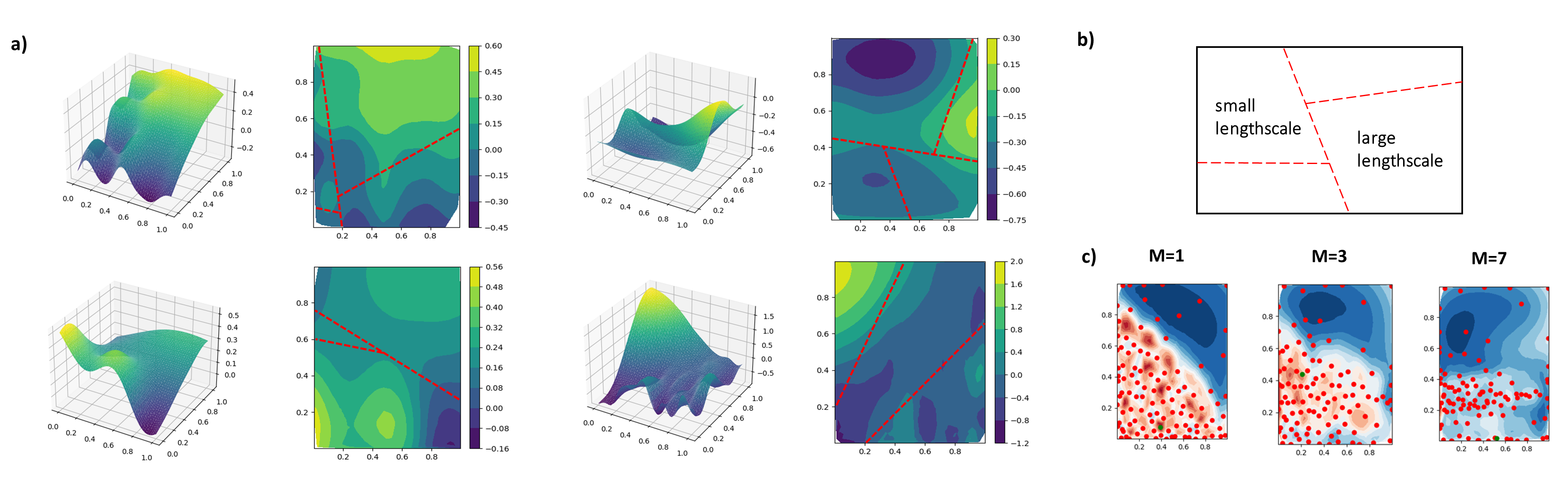}
	\vspace{.1in}
	\caption{Prior Draws and Sampling Illustrations. a) Prior draws from a GP with HHK kernel with four latent GP's (
		three hyperplanes). The sampled functions (left) and the sampled hyperplanes/partitions (right) are shown - here background colors show function values. b) Conceptual illustration of the partitioning and the different kernel parameters in different regions. c) Impact of the number of hyperplanes on the actively collected samples. Here, active learning was executed on the Exponential 2D function (details in Section \ref{section::experiments}).}
	\label{fig:priordraws}
\end{figure*}
\subsection{Kernel Parameters and Inference} 
We provide a fully-Bayesian GP model, marginalized over its hyperparameters. This has the advantage of being more robust against model misspecification in the low-data regime [see. \cite{PracticalBO,BayesianAL}].
We assume the input to be a subset of the unit square $\mathcal{X}\subset [0,1]^{d}$ and treat the kernel parameters in a Bayesian way by placing priors on them. For the hyperplanes, we introduce relevance parameters that scale the hyperplanes with $\mathbf{w}_{i}=\alpha_{i}\tilde{\mathbf{w}}_{i}$, where $\alpha_{i} \in \mathbb{R}$ and $\tilde{\mathbf{w}}_{i}\in\mathbb{R}^{d+1}$.  
Concretely, we specify the following priors on $\alpha_{i}$ and $\tilde{\mathbf{w}}_{i}$:
\begin{align*}
&\tilde{\mathbf{w}}_{i}\sim \mathcal{N}(\mathbf{0},\mathbf{I}),~~ \alpha_{i}\sim \mathrm{Gamma}(\alpha,\beta).
\end{align*}
Thus, $\mathbf{w}_{i}|\alpha_{i}\sim \mathcal{N}(0,\alpha_{i}^{2}\mathbf{I})$. This prior on $\mathbf{w}_{i}$ makes only very weak assumptions about the position of the hyperplane. However, via the prior of the relevance parameter, the scaling of the hyperplane can be influenced, which affects the slope of the sigmoid functions in (\ref{weighting_func}) and, thus, the overlap of the partitioning. The local kernels $k_{j}(x,y)$ come with their own parameters $\theta_{j}$. Our default setting utilizes RBF kernels with lengthscales and variance as parameters for which we set broad Gamma priors (see Appendix \ref{priorParam} for details). The parameters of the final model are denoted with $\Theta=\{\tilde{\mathbf{w}}_{i},\alpha_{i},\theta_{j},\sigma|i=1,\dots,M,j=1,\dots,J\}$. 

For inference, we make use of the differentiability of the kernel and employ Hamiltonian Monte Carlo (HMC) \citep{DUANE1987216}. After receiving $n$ posterior samples from $p(\Theta|\mathcal{D})$ using HMC, we employ a sample-based estimate of the marginal predictive distribution for prediction, i.e.
\begin{align*}
\label{eq:marg_dist}
p(f^{*}|x^{*},\mathcal{D})&=\int p(f^{*}|x^{*},\Theta,\mathcal{D})p(\Theta|\mathcal{D})d\Theta\\ &\approx \frac{1}{n}\sum_{\Theta_{i}\sim p(\Theta|\mathcal{D})}p(f^{*}|x^{*},\Theta_{i},\mathcal{D}).
\end{align*}
Thus, we perform a numerical sample-based approximation to the true marginalized predictive distribution.
\subsection{Induced Prior in Function Space}
In Figure \ref{fig:priordraws} a) samples from our proposed prior are shown for $M=3$ hyperplanes. The specified prior includes functions that can be described with
GP's with i) different lengthscales in different regions, ii) with different variances in different regions. Furthermore, through the sigmoidal gates the transitions between partitions is soft, leading to continuous sample functions. The partitions themselves rely on simple geometries via hyperplanes that cut the space hierarchically.

\subsection{Comparision to other Input-Partitioning Methods}
Let us emphasize some characteristics of the partitioning that distinguish our model from other input-partitioning methods like the one dimensional change-point model or the TreedGP model.

Compared to TreedGP \citep{treedGP}, the partitioning is not restricted to be axis aligned. Secondly, the partitioning is differentiable with respect to the hyperplane parameters and the inputs due to the sigmoid functions. This leads to continuous sample paths on one hand and the possibility to use gradient-based inference schemes for the hyperplanes on the other hand.

Furthermore, our partitioning can be viewed as a $d$ dimensional generalization of change-points \citep{AutomaticStatistician,NonMyopicAL}. We inherit the advantages of change-points of providing a model that can be used with gradient-based inference methods and that lead to continuous prior draws, whereas our partitioning method is applicable to $d$ dimensions. 

Concretely, our model can be formulated in a similar way to change-points [see \cite{AutomaticStatistician}] via introducing the change-hyperplane operator for two kernels $k_{1}$ and $k_{2}$ in $\mathbb{R}^{d}$
\begin{align*}
k(x,y) &= \mathcal{CH}(k_1,k_2)(x,y)\\
&=\sigma(\mathbf{w}^{\intercal}\tilde{x})\sigma(\mathbf{w}^{\intercal}\tilde{y})k_{1}(x,y) \\&+ \bar{\sigma}(\mathbf{w}^{\intercal}\tilde{x})\bar{\sigma}(\mathbf{w}^{\intercal}\tilde{y})k_{2}(x,y)
\end{align*}
with $\bar{\sigma}(\mathbf{w}^{\intercal}\tilde{x}) = 1-\sigma(\mathbf{w}^{\intercal}\tilde{x})$. The proposed kernel can be written via recursive application of the change-hyperplane operator (see Appendix \ref{section:cp_kernel}). Importantly, we note that generalizing change-points to $d$ dimensions via hyperplanes preserves a simple geometry that can be deduced with few datapoints - a necessary property for the application in active learning procedures. Additionally, the formulation via the change-hyperplane operator has the benefit that it could be utilized to dynamically search over a discrete set of trees/kernel structures, similar to what is done with the change-point operator [see \citep{AutomaticStatistician,CKS,bitzer2022structural}].
\subsection{Sample Selection}
Our goal is to provide a GP prior that is tangled to active learning settings. Thus, for the sake of simplicity we stick to the most natural way of doing active learning via querying points with highest information gain between the observation and the uncertain variables in the model \citep{ALM}. We start with an initial dataset $\mathcal{D}_{0}$ and sequentially query the oracle $f\colon\mathcal{X}\to\mathbb{R}$ at point $x_{t}$ and receive the noisy observation $y_{t}=f(x_{t})+\epsilon_{t}$ with $\epsilon_{t}\sim\mathcal{N}(0,\sigma^2)$. The next datasets are build up sequentially with $\mathcal{D}_{t}=\mathcal{D}_{t-1}\cup \{x_{t},y_{t}\}$. For the acquisition function, we use the maximum information gain between the observation $y$ and the uncertain variables which are the function $f$ and the parameters $\Theta$. Thus, the acquisition function is given by
\begin{align*}
a(x|\mathcal{D}_{t-1})&=I(y;f,\Theta|\mathcal{D}_{t-1},x)\\
&=H(y|\mathcal{D}_{t-1},x)\\&-\mathbb{E}_{p(\Theta|\mathcal{D}_{t-1})}\mathbb{E}_{p(f|\Theta,\mathcal{D}_{t-1},x)}[H(y|f,\mathcal{D}_{t-1},\Theta,x)]\\
&\propto H(y|\mathcal{D}_{t-1},x),
\end{align*}
where the right-hand term of the difference is independent of $x$ as $H(y|f,\mathcal{D}_{t-1},\Theta,x)=\mathrm{log}(\sigma \sqrt{2\pi e})$.
This acquisition function accounts for the uncertainty in the kernel parameters $\Theta$, including the uncertainty over the hyperplanes $\textbf{w}_{i},i=1,\dots,M$. In order to evaluate the acquisition function, the entropy of the marginal predictive distribution needs to be approximated, e.g. by quadrature (see Appendix \ref{quadrature}). The next query location is then obtained by 
$$
x_{t}=\mathrm{argmax}_{x\in\mathcal{X}}a(x|\mathcal{D}_{t-1}).
$$
The optimization can be performed through grid-search, random shooting or evolutionary optimizers. 

\subsection{Induced Sampling Behavior}
Partitioning models like the proposed model have the beneficial properties that they are able to distinguish regions with different levels of contained information. For example, as illustrated in Figure \ref{fig:priordraws} b), a function might be described by local GP's, where one region has large and one has small lengthscale, e.g. a function that is flat or linear in one part of the space and highly fluctuating in another. The predictive intervals thus will be larger in the region of a small lengthscale, leading to an active sampling in that region, as for example shown in Figure \ref{fig:priordraws} c). Finding the right partition with few datapoints is thus crucial, as it determines the sampling. The hierarchical partitioning provides a good trade-off between flexibility (it is not constrained to axis-aligned or orthogonal partitions) and the possibility to identify partitions with few datapoints. This is further enabled through gradient-based inference methods that lead to high quality partition samples, such as HMC. 

In Appendix \ref{theory} we also add a theorem which illustrates the sampling behavior for a simplified setting, with fixed, sharp partitions and extreme values for the lengthscales and variances of the stationary kernels.

\section{EXPERIMENTS}
\label{section::experiments}

In the following section, we will evaluate the active learning performance of our prior on three tasks which exhibit nonstationarities and compare it against the main competitors. We consider small/medium-sized tasks in terms of input-dimensionality, as this is the common application field of Gaussian processes. In the last part of this section, we also investigate the influence of the tree size and the inference scheme on the active learning performance. Furthermore, we provide code for our method.\footnote{HHK-Code \href{https://github.com/boschresearch/Hierarchical-Hyperplane-Kernels}{\texttt{https://github.com/boschresearch\linebreak /Hierarchical-Hyperplane-Kernels}}}

\paragraph{Model Setup.} We equip our method with a symmetric tree with eight latent GPs. As stationary latent GPs, we employ RBF kernels with Gamma priors for the lengthscale and the variance. For the noise variance, we utilize an exponential prior. We chose prior parameters that induce broad priors such that many functions have sufficient support. The prior parameters can be found in Appendix \ref{priorParam}. For inference, we employ a burn-in phase of 500 iterations and continue with 5000 MCMC samples that are thinned to final 100 samples.
\paragraph{Active Learning Setup.} All tasks are available as datasets with a large number of queries already executed.
The inputs are transformed to lie in the unit cube and the outputs are normalized. The optimization of the acquisition function is done with random shooting by calculating the acquisition function on a subset of the possible queries. In practice, for an oracle with continuous input domain, one might employ Latin-Hypercube samples as grid points for optimization of the acquisition function. For evaluation, we consider the RMSE curve over the selected queries on a held out test set. 

\paragraph{Compared Methods.}
We compare our method against the following methods:
\begin{enumerate}
	
	\item[\textbf{a)}] \textbf{Random:} Here, instead of selecting the queries with maximum information gain, the queries are selected randomly.
	\item[\textbf{b)}] \textbf{RBF:}  In this case, an RBF kernel is used with the same prior on the kernel parameters as for the stationary kernels in our method, also learned with HMC and queries taken with maximum information gain.
	\item[\textbf{c)}] \textbf{Warped Multi-Index GP:} This method is presented in \citet{WaMIGp} and uses a time-warped GP with type-2 maximum likelihood inference and maximum predictive variance as acquisition function. 
	\item[\textbf{d)}] \textbf{TreedGP:} The fully-Bayesian partitioning model presented in \citet{treedGP} and used for active learning of computer simulations in \citet{gpTreedAL}. We use their $tgp$ R-package and take queries via maximum information gain (the ALM criteria). 
	\item[\textbf{e)}] \textbf{DeepGP:} This approach is presented in \citet{sauer2020active} and uses DeepGPs to capture the nonstationarity. They utilize MCMC as inference and the ALC criteria \cite{ALC} as acquisition function (see Appendix \ref{DeepGPimplementation} for further details).
\end{enumerate}

\begin{figure}[t]
	\centering
	\includegraphics[width=1.0\linewidth]{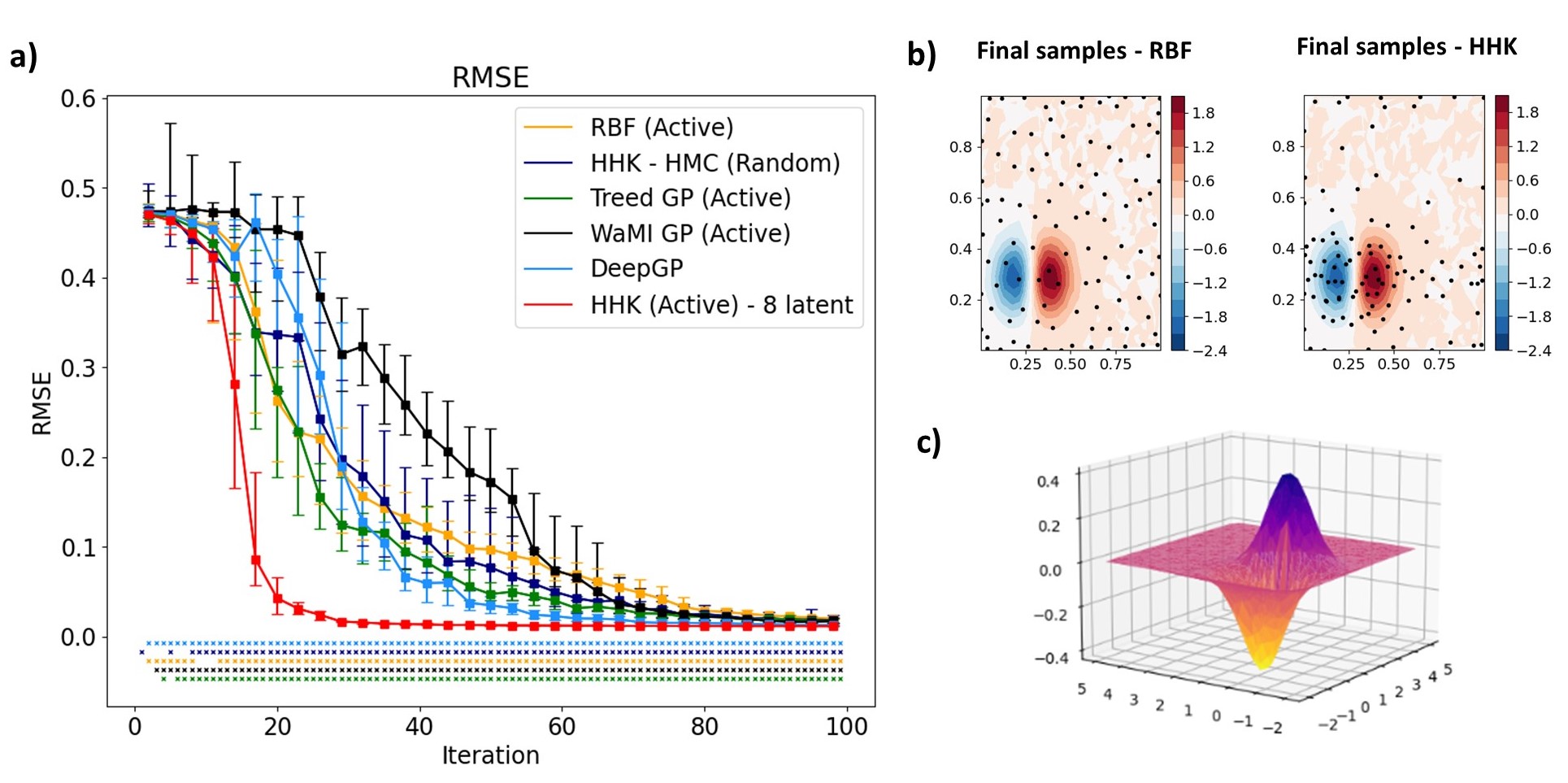}
	\vspace{0.1in}
	\caption{Exponential 2-D experiments. a) RMSE curves for the different methods. Shown are median and quartiles for each iteration over 30 runs with different initial datasets. Crosses at the bottom mark significant Wilcoxon tests in favor of our model. b) Final samples actively selected by the RBF model (left) and by our method (right) - axis are scaled to the unit cube. c) The exponential 2-D function.}
	\label{fig:summaryExp2d}
\end{figure}
\subsection{Toy-function: Exponential 2-D}

\begin{figure}[t]
	\centering
	\includegraphics[width=0.98\linewidth]{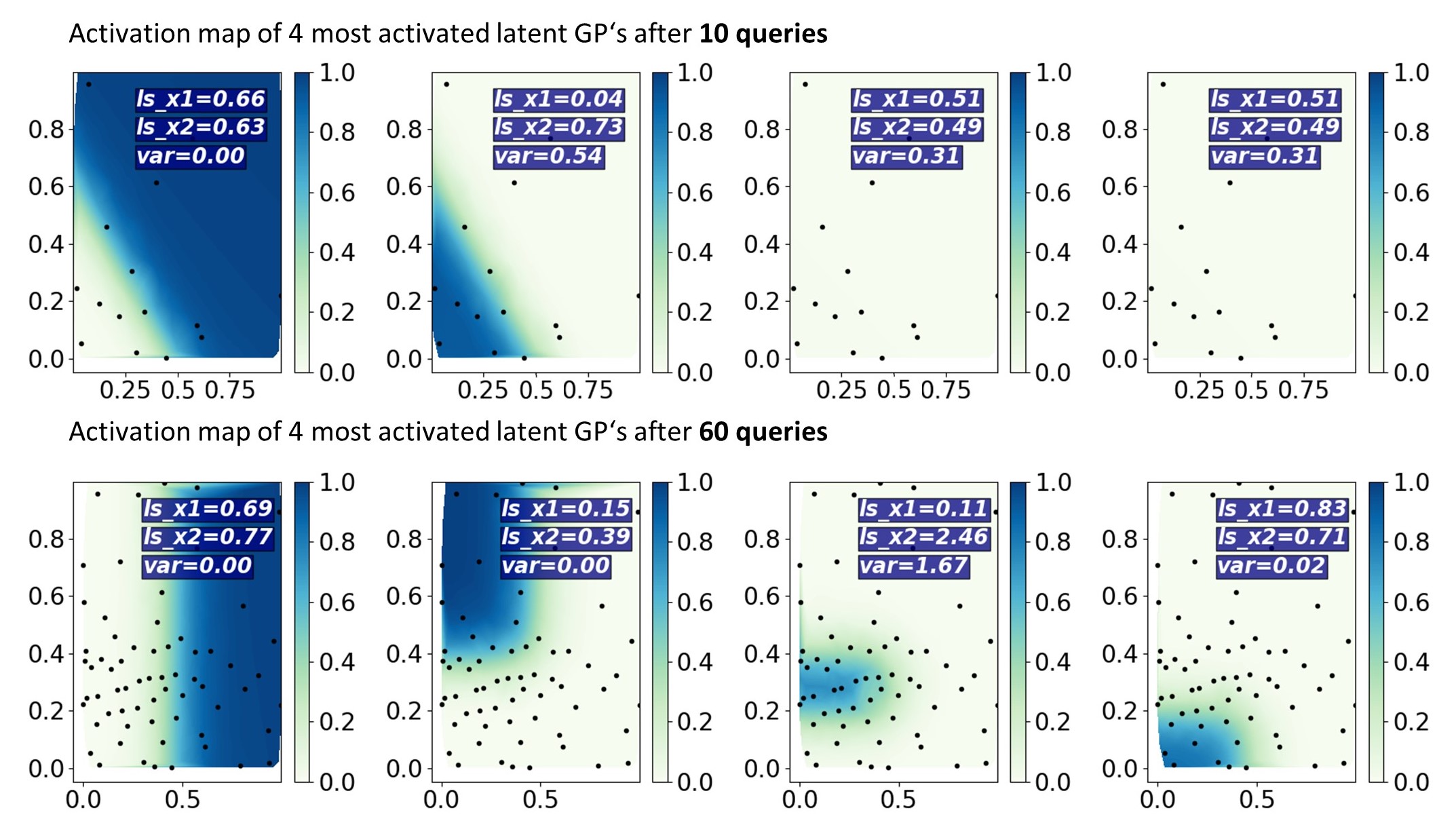}
	\vspace{.1in}
	\caption{Activation map of the partition weights (for the MAP parameters) for the four most activated stationary kernels after 10 queries (top) and 60 queries (bottom) on the Exponential 2-D dataset.}
	\label{fig:partition_summary}
\end{figure}

First, we investigate the performance of our method on the two-dimensional function
$$
f(x_{1},x_{2})=x_{1}\mathrm{exp}(-x_{1}^{2}-x_{2}^{2})
$$
with $(x_{1},x_{2})\in[-2,5]\times[-2,5]$ [see Figure \ref{fig:summaryExp2d} c)]. This test function was also investigated in \citet{gpTreedAL}. 
The function exhibits nonstationarities as it is almost flat in a large portion of the input space. Only in the lower left part of the input domain, fluctuations and larger values in the function values occur. As initial datasets we draw five data points uniformly. In comparison to the RBF model, which takes queries in a nearly space-filling manner, our method focuses  the sampling on the complex region [see Figure \ref{fig:summaryExp2d} b)]. The RMSE curves are shown in Figure \ref{fig:summaryExp2d} a) for the different methods. Besides the RBF model, our method also shows significantly better results compared to the three nonstationary methods. 
\paragraph{Impact of Partitions over the Iterations.}
We investigate the impact of the partitioning on the query selection. We show the activation maps of these weights for the four most activated kernels after 10 and 60 queries in Figure \ref{fig:partition_summary}. Here, we examine the maximum a posteriori (MAP) hyperplanes $\mathbf{\hat{w}}_{i}$ and the resulting partition weights for the corresponding stationary kernels $k_{j}$, i.e.
\begin{align*}
\lambda^{MAP}_{j}(x)=\prod_{i=1}^{M}\sigma(\mathbf{\hat{w}}_{i}^{\intercal}\tilde{x})^{\xi_{L}(j,i)}(1-\sigma(\mathbf{\hat{w}}_{i}^{\intercal}\tilde{x}))^{\xi_{R}(j,i)}.
\end{align*}
The activation weights for all other latent GPs were nearly zero. At the beginning (after 10 queries), most of the stationary kernels are turned off and the data is explained with only two latent GPs separated by a hyperplane. After 60 queries, more kernels are active, leading to a finer partitioning. The region with large kernel variance is more focused than before and almost all queries are selected in this region. We think that relying on only a few active partitions at the beginning helps to steer the samples in interesting directions. 

\subsection{Nasa - Langley-Glide-Back Booster}
\begin{figure}[t]
	\centering
	\includegraphics[width=0.93\linewidth]{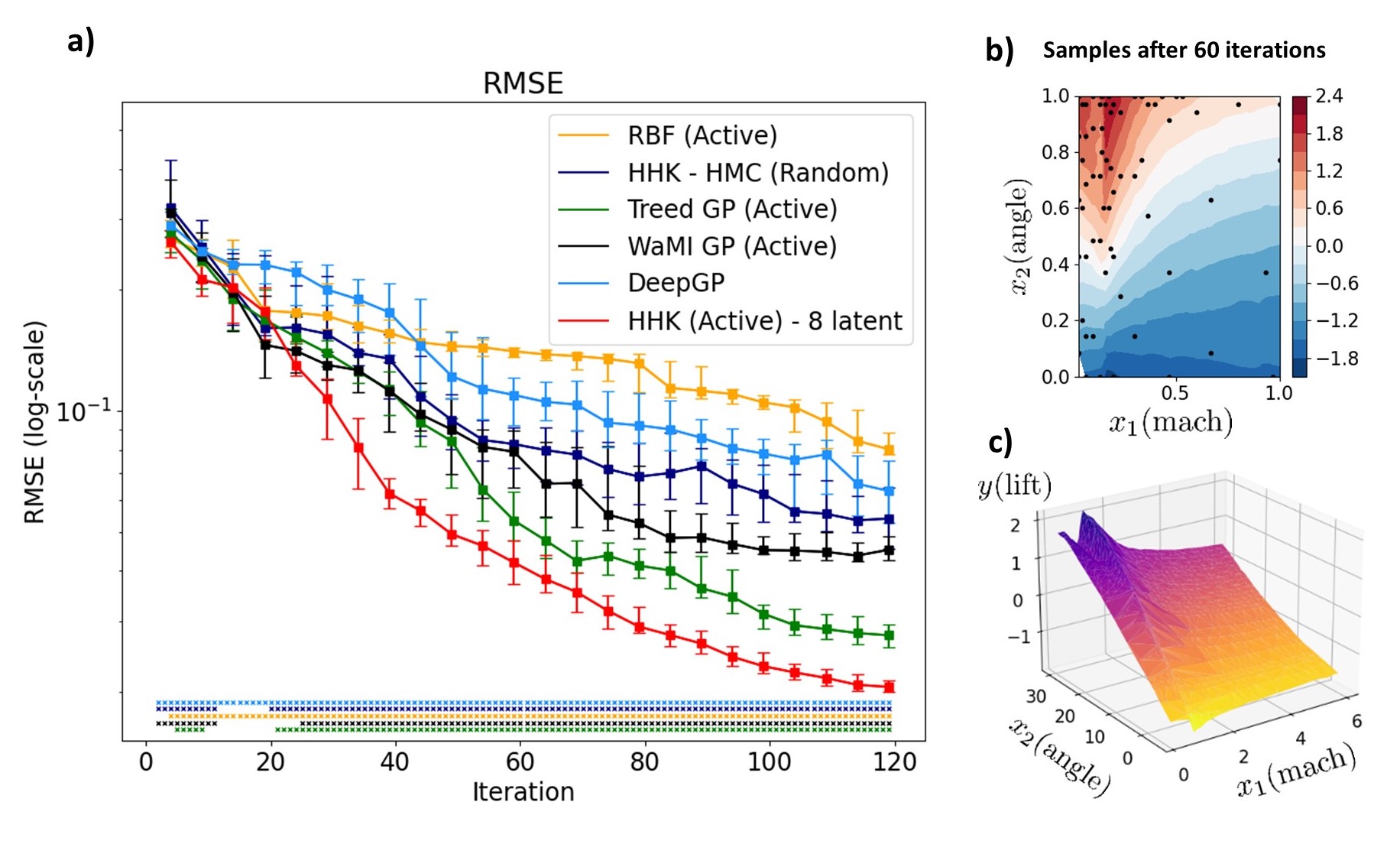}
	\vspace{.1in}
	\caption{LGBB experiments. a) RMSE curves for the different methods. The curves show medians and quartiles for each iteration over 30 runs with different initial datasets. Colored crosses at the bottom signal significant Wilcoxon tests in favor of our model against the corresponding competitor. b) Dataset after 60 queries, selected by our method. c) LGBB response surface.}
	\label{fig:summary_lgbb}
\end{figure}

The second example consists of a computer simulation (computational fluid dynamics), employed by NASA, of a rocket reentering the atmosphere, called Langley-Glide-Back Booster (LGBB). The simulation was heavily investigated for the TreedGP model in \citet{gpTreedAL} and \citet{treedGP} as it exhibits nonstationarities and has long-lasting query times. The simulation receives as inputs the angle and speed of the rocket when entering the atmosphere, and outputs the lift that the rocket exhibits. Each query to the simulation lasts several hours. Furthermore, the simulation naturally falls into two regimes, one for a speed lower than mach one and one for higher entering velocities.  Figure \ref{fig:summary_lgbb} c) shows the response surface of the simulation. The most complex region of the input space is at the upper left, where the lift value changes drastically at velocity mach one. The right part of the response surface on the other hand can be described almost by a linear function. We use the precomputed queries provided in \citet{gramacy2020surrogates} for our experiments.
We employ five uniformly drawn initial data points and make 120 active queries afterwards. Furthermore, we add an artificial noise term onto the deterministic response surface. The noise is still learned for all considered models.
Firstly, as can be seen in Figure \ref{fig:summary_lgbb} b), our method has the desired property that it focuses the samples to the complex region and only allocates very few samples to the simpler regions. This is also reflected in the predictive performance. The RMSE curves are shown in Figure \ref{fig:summary_lgbb} a). Our method leads to significantly lower RMSE values (tested with the Wilcoxon test) compared to the other methods from iteration 30 onwards. 

\subsection{Combustion Engine Noise}

\begin{figure}
	\centering
	\includegraphics[width=0.97\linewidth]{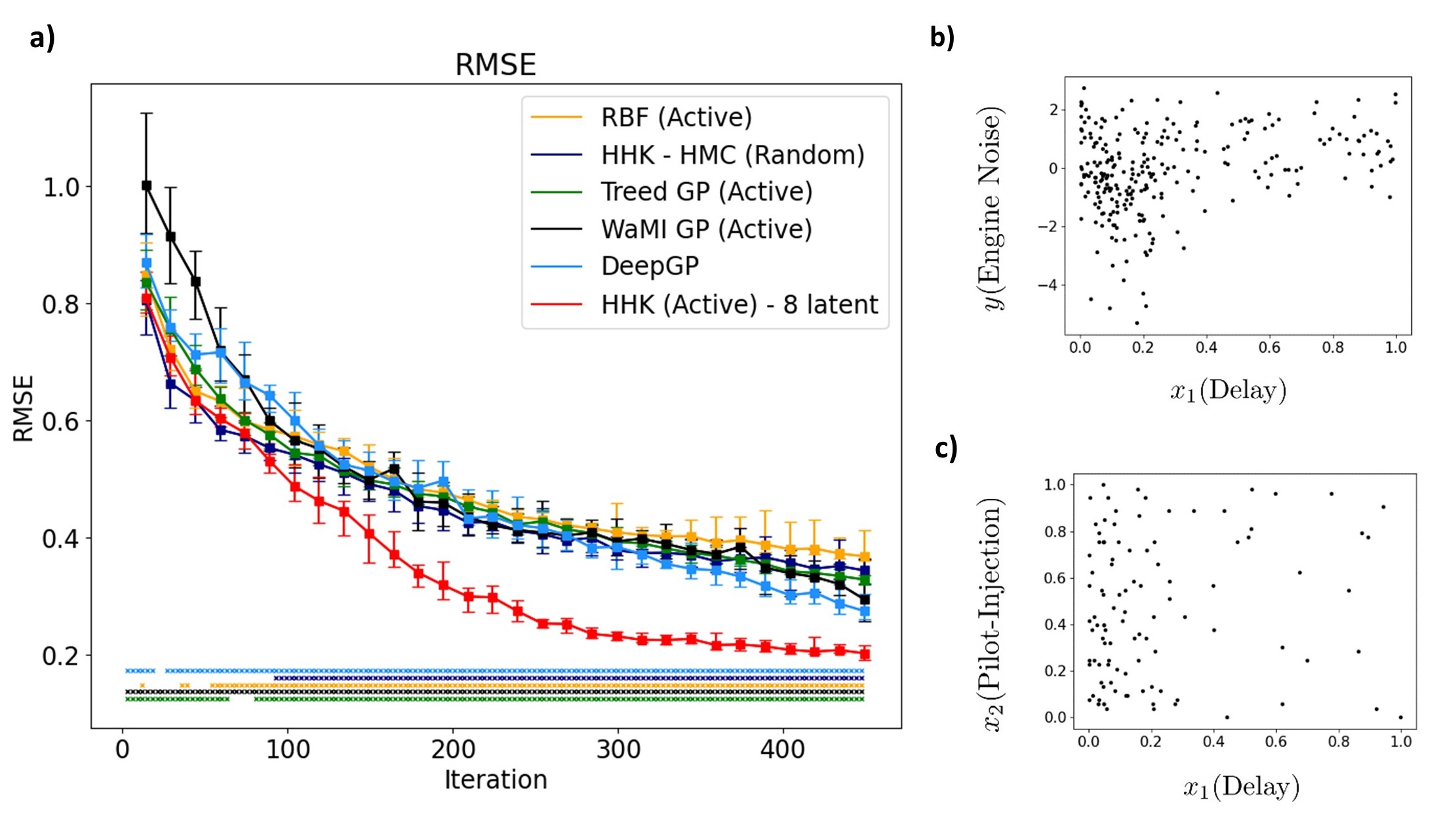}
	\vspace{0.1in}
	\caption{Combustion engine noise experiments. a) RMSE curve for the combustion engine noise dataset for the different models and querying schemes. b) Samples of $x_{1},y$ pairs to show the nonstationarity along the $x_{1}$ dimension. c) Dataset after 100 iterations actively selected by our method shown over the delay and pilot-injection dimension.}
	\label{fig:summayclosepi}
\end{figure}

In the development process of combustion engines, a significant amount of time and cost comes from the calibration of the engine control unit (ECU) [see \citet{tieze}]. To fasten the calibration process, one possibility is to use a complete simulation of the engine for a large part of the calibration procedure. Therefore, the several engine functionalities need to be simulated, which can be done in an empirical way by learning surrogate models or by physics-based models \cite{tieze}. One part of the engine calibration process is to adjust the engine noise. For that purpose, a surrogate model can be produced for the relationship between engine parameters and engine noise. The learning process bears two challenges. First, work bench time is costly and shared between several parties, and therefore the number of queries to the workbench should be kept small. Secondly, it is known that the engine noise admits major nonstationarities, in particular when varying the \textit{delay of injections} parameter [see \citet{tieze}]. The input parameters are \textit{delay of injections} $(x_{1})$, \textit{volume of pilot injection} $(x_{2})$, \textit{rail pressure} $(x_{3})$, \textit{air mass} $(x_{4})$, \textit{boost  pressure} $(x_{5})$ and \textit{controlstart (in ms)} $(x_{6})$ [see \citet{tieze}]. The output is the \textit{engine noise} in dB. 
We start with an initial randomly drawn dataset of size five. The resulting RMSE curves can be seen in Figure \ref{fig:summayclosepi} a). As with the other two experiments, our method provides more precise predictions with fewer selected data points.
Also, as expected, the model queries more data points for lower values of $x_{1}$ as can be seen in Figure \ref{fig:summayclosepi} c).       
\subsection{Influence of the Tree Size}
\begin{figure}[t]
	\centering
	\includegraphics[width=0.99\linewidth]{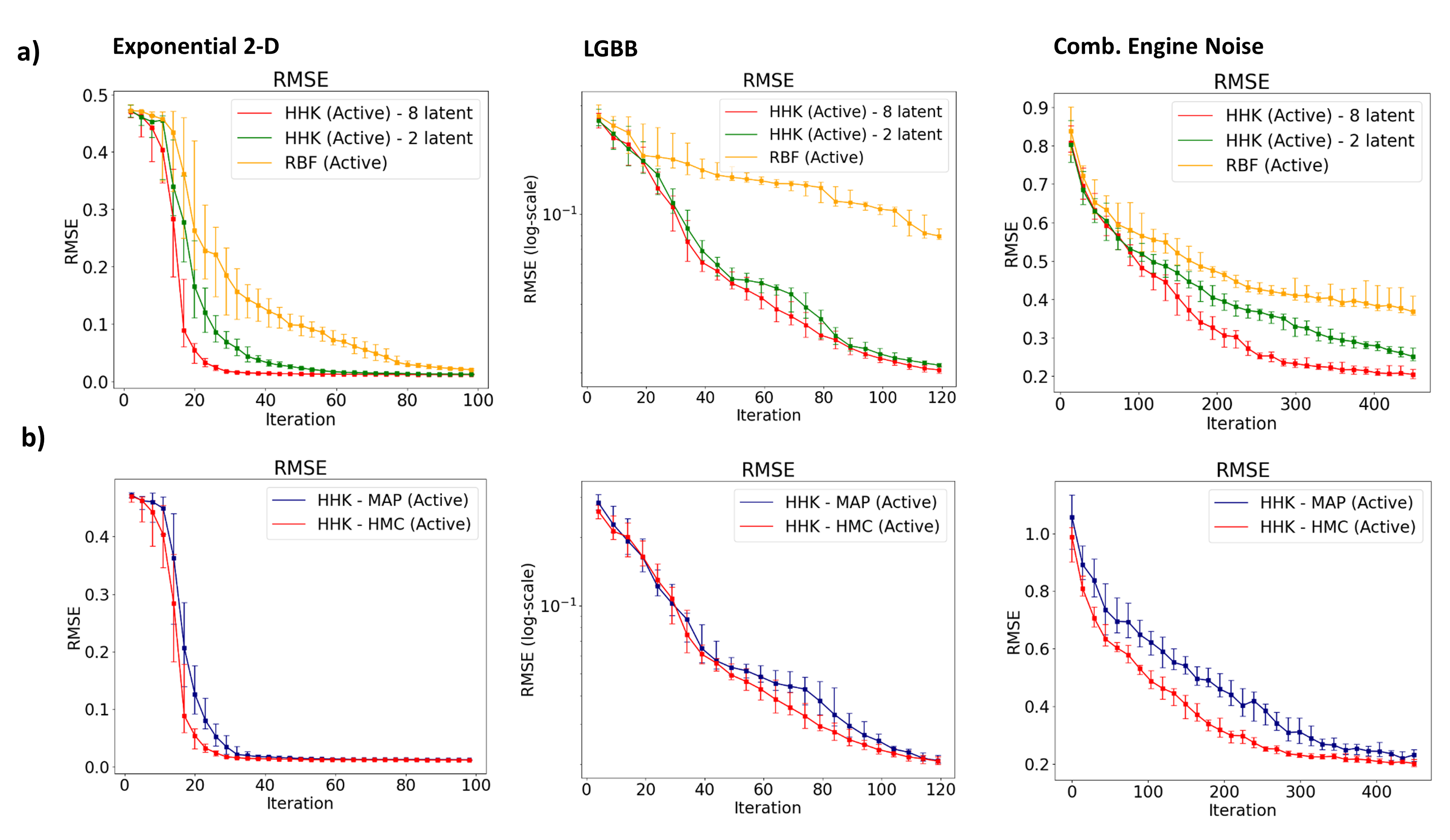}
	\vspace{0.1in}
	\caption{a) RMSE curves for different tree sizes (1 (stat. RBF), 2 and 8 latent GPs) over the three datasets. b) RMSE curves for HMC vs MAP inference for a HHK
		with 8 latent GPs.}
	\label{fig:treeandinferencesummary}
\end{figure}

We examine the impact of the tree size on the performance of the active learning procedure. Figure \ref{fig:treeandinferencesummary} a) shows the RMSE curves for the trivial tree (stationary RBF kernel), a HHK model with two and with eight latent GPs on all considered datasets.
We see that for the Exponential 2-D data and the combustion engine noise dataset, the active learning procedure benefits from larger trees. For the LGBB dataset, we can see that even adding one hyperplane seems to give a large performance gain. This is plausible since the CFD simulation has two regimes that can be distinguished with a hyperplane. For the combustion engine noise data, previous expert experience [see \citet{tieze}] has shown that nonstationarity occurs mainly within one dimension. However, we find that we gain performance by using finer partitioning with more hyperplanes [see Figure \ref{fig:treeandinferencesummary}].
\subsection{Influence of the Inference Scheme}
In Figure \ref{fig:treeandinferencesummary} b), we investigate the influence of the kernel parameter inference on the active learning performance. To this end, we compare HMC inference with optimization of the GP hyperparameters using maximum-a-posteriori (MAP). We observe that using a Bayesian approach via HMC leads to a faster convergence compared to MAP on the higher-dimensional ClosePI task. On the other two tasks we observe a slight advantage of HMC. This also supports the recent finding of \cite{BayesianAL} that marginalizing over kernel parameters is beneficial for active learning. However, we note that on LGBB and Exponential 2-D the difference between HMC and MAP ist not very large, and MAP might be a reasonable choice in cases where, for example, inference time is important (see Appendix \ref{inferencetime} for details on inference time). 

%We think in particular for nonstationary kernels misspecified hyperparameters, resulting from point-estimates on very small datasets, might guide the sampling into wrong directions.

\section{CONCLUSION}
In this work, we presented a new input-partitioning kernel for GP regression and investigated its active learning performance on a variety of tasks. Our method induces partitioning via a hierarchy of hyperplanes, having the advantage of preserving useful properties such as differentiability of the kernel and smooth sample functions, while keeping a simple geometry. We show that the induced active sampling focuses on the complicated region of the input space, and that our method significantly outperforms previous methods, including previous partitioning models, on real-world active learning tasks.

\bibliographystyle{abbrvnat}
\bibliography{lib}{}

% If you have textual supplementary material
\appendix
\onecolumn

\section{METHODOLOGICAL CONSIDERATIONS}

\subsection{Formulation via Change-Hyperplanes}
\label{section:cp_kernel}
In \cite{AutomaticStatistician} the change-point operator is introduced for two kernels $k_{1}(x,y)$ and $k_{2}(x,y)$ and a point $s\in\mathbb{R}$
$$
\mathcal{CP}_{s}(k_{1},k_{2})(x,y)=\sigma(x-s)\sigma(y-s)k_{1}(x,y)+\bar{\sigma}(x-s)\bar{\sigma}(y-s)k_{2}(x,y)
$$
with $\bar{\sigma}(x-s)=1-\sigma(x-s)$ and $x,y\in \mathbb{R}$. Multiple change-points can be applied via iterative application of this operator. For example, for three change-points $s_{1},s_{2},s_{3} \in \mathbb{R}$ the final kernel can be defined as
$$
k(x,y)=\mathcal{CP}_{s_{1}}(\mathcal{CP}_{s_{2}}(k_{1},k_{2}),\mathcal{CP}_{s_{3}}(k_{3},k_{4})),
$$
In case $k_{1},\dots,k_{4}$ are stationary, the final kernel would induce functions that behave stationary in-between the change-points, but are less correlated/behave differently between two separate intervals.

The change-point operator is defined on one-dimensional inputs, e.g. $s,x,y\in\mathbb{R}$. Our kernel can be viewed as a $d$ dimensional generalization of change-points. We illustrate this for the example in Figure 1 (in the main paper) with a symmetric tree with three hyperplanes. We denote $\tilde{x}=(1,x)^{\intercal}\in\mathbb{R}^{d+1}$ and define the change-hyperplane operator as 
\begin{align*}
\mathcal{CH}_{\mathbf{w}}(k_1,k_2)(x,y) &=\sigma(\mathbf{w}^{\intercal}\tilde{x})\sigma(\mathbf{w}^{\intercal}\tilde{y})k_{1}(x,y)+ \bar{\sigma}(\mathbf{w}^{\intercal}\tilde{x})\bar{\sigma}(\mathbf{w}^{\intercal}\tilde{y})k_{2}(x,y),~~~x,y\in\mathbb{R}^{d},
\end{align*}
where $\bar{\sigma}(\mathbf{w}^{\intercal}\tilde{x}) = 1-\sigma(\mathbf{w}^{\intercal}\tilde{x})$. 
For the HHK with three hyperplanes [Figure 1 a)], the HHK weights $\lambda_{j}(x)$ result to:
\begin{align*}
\lambda_{1}(x)&=\prod_{i=1}^{3}\sigma(\mathbf{w}_{i}^{\intercal}\tilde{x})^{\xi_{L}(1,i)}(1-\sigma(\mathbf{w}_{i}^{\intercal}\tilde{x}))^{\xi_{R}(1,i)}=\sigma(\mathbf{w}_{1}^{\intercal}\tilde{x})\sigma(\mathbf{w}_{2}^{\intercal}\tilde{x}),\\
\lambda_{2}(x)&=\prod_{i=1}^{3}\sigma(\mathbf{w}_{i}^{\intercal}\tilde{x})^{\xi_{L}(2,i)}(1-\sigma(\mathbf{w}_{i}^{\intercal}\tilde{x}))^{\xi_{R}(2,i)}=\sigma(\mathbf{w}_{1}^{\intercal}\tilde{x})(1-\sigma(\mathbf{w}_{2}^{\intercal}\tilde{x}))=\sigma(\mathbf{w}_{1}^{\intercal}\tilde{x})\bar{\sigma}(\mathbf{w}_{2}^{\intercal}\tilde{x}),\\
\lambda_{3}(x)&=\prod_{i=1}^{3}\sigma(\mathbf{w}_{i}^{\intercal}\tilde{x})^{\xi_{L}(3,i)}(1-\sigma(\mathbf{w}_{i}^{\intercal}\tilde{x}))^{\xi_{R}(3,i)}=(1-\sigma(\mathbf{w}_{1}^{\intercal}\tilde{x}))\sigma(\mathbf{w}_{3}^{\intercal}\tilde{x})=\bar{\sigma}(\mathbf{w}_{1}^{\intercal}\tilde{x})\sigma(\mathbf{w}_{3}^{\intercal}\tilde{x}),\\
\lambda_{4}(x)&=\prod_{i=1}^{3}\sigma(\mathbf{w}_{i}^{\intercal}\tilde{x})^{\xi_{L}(4,i)}(1-\sigma(\mathbf{w}_{i}^{\intercal}\tilde{x}))^{\xi_{R}(4,i)}=(1-\sigma(\mathbf{w}_{1}^{\intercal}\tilde{x}))(1-\sigma(\mathbf{w}_{3}^{\intercal}\tilde{x}))=\bar{\sigma}(\mathbf{w}_{1}^{\intercal}\tilde{x})\bar{\sigma}(\mathbf{w}_{3}^{\intercal}\tilde{x}),
\end{align*}
Thus, the kernel can be written via the change-hyperplane operator as
\begin{align*}
k(x,y)&=\sum_{j=1}^{4}\lambda_{j}(x)\lambda_{j}(y)k_{j}(x,y)\\ &=\sigma(\mathbf{w}_{1}^{\intercal}\tilde{x})\sigma(\mathbf{w}_{2}^{\intercal}\tilde{x})\sigma(\mathbf{w}_{1}^{\intercal}\tilde{y})\sigma(\mathbf{w}_{2}^{\intercal}\tilde{y}) k_{1}(x,y)+\sigma(\mathbf{w}_{1}^{\intercal}\tilde{x})\bar{\sigma}(\mathbf{w}_{2}^{\intercal}\tilde{x})\sigma(\mathbf{w}_{1}^{\intercal}\tilde{y})\bar{\sigma}(\mathbf{w}_{2}^{\intercal}\tilde{y})k_{2}(x,y)\\
&+\bar{\sigma}(\mathbf{w}_{1}^{\intercal}\tilde{x})\sigma(\mathbf{w}_{3}^{\intercal}\tilde{x}) \bar{\sigma}(\mathbf{w}_{1}^{\intercal}\tilde{y})\sigma(\mathbf{w}_{3}^{\intercal}\tilde{y})k_{3}(x,y)+\bar{\sigma}(\mathbf{w}_{1}^{\intercal}\tilde{x})\bar{\sigma}(\mathbf{w}_{3}^{\intercal}\tilde{x})\bar{\sigma}(\mathbf{w}_{1}^{\intercal}\tilde{y})\bar{\sigma}(\mathbf{w}_{3}^{\intercal}\tilde{y})k_{4}(x,y)\\
&=\sigma(\mathbf{w}_{1}^{\intercal}\tilde{x})\sigma(\mathbf{w}_{1}^{\intercal}\tilde{y})\bigg(\sigma(\mathbf{w}_{2}^{\intercal}\tilde{x})\sigma(\mathbf{w}_{2}^{\intercal}\tilde{y}) k_{1}(x,y)+\bar{\sigma}(\mathbf{w}_{2}^{\intercal}\tilde{x})\bar{\sigma}(\mathbf{w}_{2}^{\intercal}\tilde{y})k_{2}(x,y)\bigg)\\
&+\bar{\sigma}(\mathbf{w}_{1}^{\intercal}\tilde{x})\bar{\sigma}(\mathbf{w}_{1}^{\intercal}\tilde{y})\bigg(\sigma(\mathbf{w}_{3}^{\intercal}\tilde{x})\sigma(\mathbf{w}_{3}^{\intercal}\tilde{y}) k_{3}(x,y)+\bar{\sigma}(\mathbf{w}_{3}^{\intercal}\tilde{x})\bar{\sigma}(\mathbf{w}_{3}^{\intercal}\tilde{y})k_{4}(x,y)\bigg)\\
&=\sigma(\mathbf{w}_{1}^{\intercal}\tilde{x})\sigma(\mathbf{w}_{1}^{\intercal}\tilde{y}) \mathcal{CH}_{\mathbf{w_{2}}}(k_1,k_2)(x,y) +\bar{\sigma}(\mathbf{w}_{1}^{\intercal}\tilde{x})\bar{\sigma}(\mathbf{w}_{1}^{\intercal}\tilde{y})\mathcal{CH}_{\mathbf{w_{3}}}(k_3,k_4)(x,y)\\
&=\mathcal{CH}_{\mathbf{w_{1}}}(\mathcal{CH}_{\mathbf{w_{2}}}(k_1,k_2),\mathcal{CH}_{\mathbf{w_{3}}}(k_3,k_4))(x,y)
\end{align*}
Thus, one might interpret the HHK as iterative application of the change-hyperplane operator along the nodes of its tree. 

\newtheorem{theorem}{Theorem}
\newtheorem{lemma}{Lemma}
\subsection{Theory: Sampling Behavior of Partitioning Models}
\label{theory}
In the following subsection, $f$ is drawn from a Gaussian process with kernel $k$, $A =\{x_{1},\dots,x_{T}\} \subset D$ is a set of input locations with $D\subset \mathcal{X}$ compact, $f_{A}=[f(x)]_{x\in A}$, and $y_{A}=f_{A}+\epsilon_{A}$ is the resulting observation with noise $\epsilon_{A}\sim \mathcal{N}(0,\sigma^{2}\mathbf{I})$. The Shannon information $I(\cdot)$ between observations and latent function can be calculated in closed form with
$
I(y_{A},f|k)=I(y_{A},f_{A}|k)=\frac{1}{2}\mathrm{log}|\mathbf{I}+\sigma^{-2}K_{A}|
$,
where $K_{A}=[k(x,y)]_{x,y\in A}$ is the Gram matrix at locations $A$.

We want to gain theoretical insights into the sampling behavior that is induced by partitioning kernels. For that, we take a detailed look at the sampling behavior in some extreme cases of the kernel parameters. We consider the general partitioning kernel $k(x,y)=\sum_{j=1}^{J} \mathbf{1}_{\{x,y\in D_{j}\}}k^{l_{j},\sigma_{j}^{2}}_{j}(x,y)$ for some partition  $D_{j},j=1,\dots,J$, of the input space. Here, $k^{l_{j},\sigma_{j}^{2}}_{j}, j=1,\dots,J$ are stationary kernels with lengthscales $l_{j}$ and variances $\sigma_{j}^{2}$, for example $k^{l_{j},\sigma_{j}^{2}}_{j}=\sigma_{j}^{2}\mathrm{exp}\bigg(-\frac{\Vert x-y\Vert_{2}^{2}}{l_{j}^{2}}\bigg)$. We consider a simplified setting in the sense that the partitions are sharp (defined via indicator functions) and furthermore the kernel parameters, including the partitions, are considered \textit{fixed}. Our kernel becomes this form in the limit of the smoothness parameter in the sigmoid function (see Lemma \ref{lemma5}). Furthermore, the MAP estimate of the TreedGP model would be a special case of this kernel family (that has axis-aligned partitions $D_{j}$). We note that in case of fixed kernel parameters the maximum information gain acquisition function simplifies to
\begin{equation}
\label{selection_fixed2}
a(x|\mathcal{D}_{t-1})=I(y;f|\mathcal{D}_{t-1},x,\Theta) \propto \sigma_{t-1}^{2}(x|\Theta)
\end{equation}
where $\sigma_{t-1}^{2}(x|\Theta)$ is the predictive variance of the posterior GP $f|\mathcal{D}_{t-1}$ with parameters $\Theta$. The acquisition function (\ref{selection_fixed2}) is the greedy step to maximize the complete information $I(y_{A},f_{A}|k)$ [see \citet{NoRegretAndED}]. We will therefore analyze the samples $A$ that maximize $I(y_{A},f_{A}|k)$ with $|A|=T$ and $T\in \mathbb{N}$. In Theorem \ref{theorm1} we take a detailed look at the optimally selected samples, when the kernel exhibits the extreme cases, where either the variance of one local kernel is zero or its lengthscale goes to infinity.

The theorem states that, when the partition is sharp, the sampling will ignore the region, where the kernel variance is zero and will only need a tiny fraction of the region, where the kernel lengthscale is very large. A look into the proof reveals that, in the regions that exhibit large lengthscales, the function can be considered almost as constant, such that datapoints from a tiny portion of this area are sufficient to deduce the function values in the remaining part. In the regions exhibiting zero variance, there is no information left, such that no datapoints need to be gathered here. Instead, the samples can be allocated to regions where more information is present. In practice, the variance will not be zero and the lengthscale might have very large, but finite values. In the experimental section of the main paper we see examples for input regions that exhibit very small variance or large lengthscales, and we observe the indicated sampling behavior. 

It is important to note that, in practice, the correct identification of the different partitions is crucial for the sampling behavior to behave as described. This is where the properties of our kernel play an important role, as the hierarchical hyperplanes allow for flexible partitions that can be learned efficiently due to the differentiability of the kernel.
\begin{theorem}
	\label{theorm1}
	Let $D\subset\mathbb{R}^{d}$ be compact and $D_{j}\subset D, j=1,\dots,J$, with non-empty interior and such that $D_{i}\cap D_{j} = \emptyset$ and $\bigcup_{j=1}^{J} D_{j} = D$. Let $k(x,y)=\sum_{j=1}^{J} \mathbf{1}_{\{x,y\in D_{j}\}}k^{l_{j},\sigma_{j}^{2}}_{j}(x,y)$, where $k^{l_{j},\sigma_{j}^{2}}_{j}, j=1,\dots,J$ are stationary kernels with lengthscales $l_{j}$ and variances $\sigma_{j}^{2}$, e.g. $k^{l_{j},\sigma_{j}^{2}}_{j}=\sigma_{j}^{2}\mathrm{exp}\bigg(-\frac{\Vert x-y\Vert_{2}^{2}}{l_{j}^{2}}\bigg)$. Consider the two cases that for some $i \in \{1,\dots,J\}$ either 1) the variance is $\sigma_{i}^{2}=0$ or 2) the lengthscale $l_{i}\to\infty$. Denote in both cases the resulting kernel on the full input space with $k^{*}(x,y)$ (which is the p.w. limit for case 2). Then it holds:\\
	\vspace{-0.3cm}
	
	For case 1): 
	\begin{align*}
	&\exists X^{*}=\{x_{1},\dots,x_{T}\} \subset D \setminus D_{i}:\\ &X^{*} \in \underset{A\subset D , |A|=T}{\mathrm{argmax}} I(y_{A},f_{A}|k^{*}).
	\end{align*}
	
	For case 2): Let $\xi$ be an arbitrary interior point in $D_{i}$. Then for some $a>0$ and any $0<\epsilon<a$:
	\begin{align*}
	&\exists X^{*}=\{x_{1},\dots,x_{T}\} \subset (D \setminus D_{i})\cup B_{\epsilon}(\xi): \\&X^{*} \in \underset{A\subset D , |A|=T}{\mathrm{argmax}} I(y_{A},f_{A}|k^{*}),
	\end{align*}
	where $B_{\epsilon}(\xi)$ is the open ball around $\xi$ with radius $\epsilon$.
	
\end{theorem}

\textbf{Proof.} Let $A\subset D$ with $|A|=T$. Now, let $A_{j}:=A\cap D_{j}$. Then $\bigcup_{j=1}^{J}A_{j}=A$ and $A_{j}\cap A_{j^{'}}=\emptyset$ for all $j\neq j^{'}$. As $k(x,y)=0$ for $x \in D_{j}$ and $y \in D_{j^{'}}$ with $j\neq j^{'}$, it holds that $y_{A_{j}}, j=1,\dots,J$ are independent from each other, hence,
$$
H(y_{A})=\sum_{j=1}^{J}H(y_{A_{j}}).
$$
Additionally, $y_{A_{j}}|f_{A_{j}}, j=1,\dots,J$ are independent (as the noise term is i.i.d.) and, thus,
$$ 
H(y_{A}|f_{A})= \sum_{j=1}^{J}H(y_{A_j}|f_{A_{j}}).
$$ 
We denote $K^{j}_{A}:=[k_{j}(x,y)]_{x,y\in A}$ and note that $K_{A_{j}}=K^{j}_{A_{j}}$ for all $j=1,\dots,J$ as $A_{j}\subset D_{j}$. Therefore, it holds,
\begin{align*}
I(y_{A},f_{A}|k)&=H(y_{A})-H(y_{A}|f_{A})=\sum_{j=1}^{J}H(y_{A_j})-\sum_{j=1}^{J}H(y_{A_j}|f_{A_{j}})\\
&=\sum_{j=1}^{J} I(y_{A_{j}},f_{A_{j}}|k)=\sum_{j=1}^{J}\frac{1}{2}\mathrm{log}|\mathbf{I}+\sigma^{-2}K_{A_{j}}|=\sum_{j=1}^{J}\frac{1}{2}\mathrm{log}|\mathbf{I}+\sigma^{-2}K_{A_{j}}^{j}|\\
& = \sum_{j=1}^{J} I(y_{A_{j}},f_{A_{j}}|k_{j}). 
\end{align*}
Case 1): For this case, where $\sigma_{i}^{2}=0$ for some $i \in \{1,\dots,J\}$, it additionally holds,
$$
I(y_{A},f_{A}|k^{*})= \sum_{j=1}^{J} I(y_{A_{j}},f_{A_{j}}|k_{j}) = \sum_{j\neq i} I(y_{A_{j}},f_{A_{j}}|k_{j}),
$$
because $k_{i}(x,y)=0$ is the trivial kernel, since $\sigma_{i}^{2}=0$, and it holds
$$
I(y_{A_{i}},f_{A_{i}}|k_{i})=\frac{1}{2}\mathrm{log}|\mathbf{I}+\sigma^{-2}K_{A_{i}}^{i}|=\frac{1}{2}\mathrm{log}|\mathbf{I}|=\frac{1}{2}\mathrm{log}|1|=0.
$$ 
This yields
\begin{align*}
\underset{A\subset D , |A|=T}{\mathrm{max}} I(y_{A},f_{A}|k^{*})&=\underset{A\subset D , |A|=T}{\mathrm{max}} \sum_{j=1}^{J} I(y_{A \cap D_{j}},f_{A \cap D_{j}}|k_{j})=\underset{A\subset D , |A|=T}{\mathrm{max}} \sum_{j\neq i} I(y_{A \cap D_{j}},f_{A \cap D_{j}}|k_{j})\\
&=\underset{A\subset D\setminus D_{i} , |A|=T}{\mathrm{max}} \sum_{j\neq i} I(y_{A \cap D_{j}},f_{A \cap D_{j}}|k_{j})=\underset{A\subset D\setminus D_{i} , |A|=T}{\mathrm{max}} I(y_{A},f_{A}|k^{*}).
\end{align*}
\\
Case 2): For the second case, where $\l_{i}\to \infty$ for some $i\in\{1,\dots,J\}$, we denote the pointwise (p.w.) limit of $k_{i}(x,y)$ with $k^{*}_{i}(x,y)$ and it holds that $k^{*}_{i}(x,y)=c$ with some constant $c>0$ (actually it holds that $c=\sigma^{2}_{i}$). We denote $K_{A}^{*,i}:=[k^{*}_{i}(x,y)]_{x,y\in A}$. Let $\xi$ be an interior point in $D_{i}$ and $a>0$ such that $B_{a}(\xi)\subset D_{i}$. Then, for any $\tilde{A}_{i}\subset B_{\epsilon}(\xi)$ with $|\tilde{A}_{i}|=|A_{i}|$ and $0<\epsilon<a$ it holds that $K_{A_{i}}^{*,i}=K_{c}=K_{\tilde{A}_{i}}^{*,i}$ with $K_{c}:=[c]_{x,y\in A_{i}}$ and thus
$$
I(y_{A_{i}},f_{A_{i}}|k^{*}_{i})=\frac{1}{2}\mathrm{log}|\mathbf{I}+\sigma^{-2}K_{c}|=I(y_{\tilde{A}_{i}},f_{\tilde{A}_{i}}|k^{*}_{i}).
$$ 
Thus, we obtain
\begin{align*}
\underset{A\subset D , |A|=T}{\mathrm{max}} I(y_{A},f_{A}|k^{*})&=\underset{A\subset D , |A|=T}{\mathrm{max}} \bigg(I(y_{A\cap D_{i}},f_{A\cap D_{i}}|k^{*}_{i})+ \sum_{j\neq i} I(y_{A \cap D_{j}},f_{A \cap D_{j}}|k_{j}) \bigg)\\
&=\underset{A\subset (D \setminus D_{i})\cup B_{\epsilon}(\xi) , |A|=T}{\mathrm{max}} \bigg(I(y_{A\cap B_{\epsilon}(\xi)},f_{A\cap B_{\epsilon}(\xi)}|k^{*}_{i})+ \sum_{j\neq i} I(y_{A \cap D_{j}},f_{A \cap D_{j}}|k_{j}) \bigg)\\&=\underset{A\subset  (D \setminus D_{i})\cup B_{\epsilon}(\xi) , |A|=T}{\mathrm{max}} I(y_{A},f_{A}|k^{*}).~~~~~~\blacksquare
\end{align*}

The following lemma shows that the Hierarchical-Hyperplane Kernel reaches the considered form $k(x,y)=\sum_{j=1}^{J} \mathbf{1}_{\{x,y\in D_{j}\}}k^{l_{j},\sigma_{j}^{2}}_{j}(x,y)$ for some partition  $D_{j},j=1,\dots,J$ in the limit of the smoothness parameter. Furthermore, the proof reveals the exact form of the sharp partitions $D_{j}$.
\begin{lemma}
	\label{lemma5}
	Let $k(x,y|[\mathbf{w}_{i}])=\sum_{j=1}^{J}\lambda_{j}(x,[\mathbf{w}_{i}])\lambda_{j}(y,[\mathbf{w}_{i}])k_{j}(x,y)$ be the Hierarchical Hyperplane Kernel (HHK), thus $\lambda_{j}(x,[\mathbf{w}_{i}])=\prod_{i=1}^{M}\sigma(\mathbf{w}_{i}^{\intercal}\tilde{x})^{\xi_{L}(j,i)}(1-\sigma(\mathbf{w}_{i}^{\intercal}\tilde{x}))^{\xi_{R}(j,i)}$ with $\tilde{x}=(1,x)^{\intercal}$ and $k_{j}$ be kernels on $\mathbb{R}^{d}$. Then there exists sequences $(\mathbf{w}_{i}^{(n)})_{n\ge0}$ with $\Vert\mathbf{w}_{i}^{(n)} \Vert \to \infty$, for all $i=1,\dots,M$, and $D_{j}\subset \mathbb{R}^{d}, j=1,\dots,J$ such that
	$$
	k(x,y|[\mathbf{w}^{(n)}_{i}])\to \sum_{j=1}^{J} \mathbf{1}_{\{x,y\in D_{j}\}}k_{j}(x,y)
	$$
	pointwise.% where each set $S_{i}:=\{x\in \mathbb{R}^{d}|\mathbf{w}_{i}^{\intercal}\tilde{x}=0\}$ is part of the boundary of $D_{j}$ for some $j\in\{1,\dots,J\}$.\\\\
\end{lemma}
\textbf{Proof.} Let $\alpha_{n}=n^{2}$, $\beta_{n}=\frac{1}{n}$ and
$$
\mathbf{w}_{i}^{(n)}:=\alpha_{n}\bigg(\tilde{\mathbf{w}}_{i}+\beta_{n}\left(
\begin{array}{c}
1\\
0\\
\vdots\\
0
\end{array}
\right)\bigg)
$$
for some fixed $\tilde{\mathbf{w}}_{i}, i=1,\dots,M$.
Then, for all $i=1,\dots,M$, it holds
$$
\sigma(\mathbf{w}_{i}^{(n)\intercal}\tilde{x})
=\frac{1}{1+\mathrm{exp}(-\alpha_{n}(\tilde{\mathbf{w}}_{i}^{\intercal}\tilde{x}+\beta_{n}))}\to \begin{cases}
1~~~\text{for}~~~\tilde{\mathbf{w}}_{i}^{\intercal}\tilde{x}\ge 0,\\
0~~~\text{for}~~~\tilde{\mathbf{w}}_{i}^{\intercal}\tilde{x}< 0.
\end{cases}
$$
Therefore, we obtain\\
\begin{align*}
&\lambda_{j}(x,[\mathbf{w}^{(n)}_{i}])=\prod_{i=1}^{M}\sigma(\mathbf{w}_{i}^{(n)\intercal}\tilde{x})^{\xi_{L}(j,i)}(1-\sigma(\mathbf{w}_{i}^{(n)\intercal}\tilde{x}))^{\xi_{R}(j,i)}\\
&\to \begin{cases}
1~~~\text{if for all}~~i~~\text{with}~~\xi_{L}(j,i)=1:  \tilde{\mathbf{w}}_{i}^{\intercal}\tilde{x}\ge 0~~\text{and all}~~i~~\text{with}~~\xi_{R}(j,i)=1: \tilde{\mathbf{w}}_{i}^{\intercal}\tilde{x}< 0, \\0~~~\text{else}.
\end{cases}
\end{align*}
We set 
$$
D_{j}:=\{x\in\mathbb{R}^{d}:\lambda_{j}(x,[\mathbf{w}^{(n)}_{i}])\to 1\}
$$
so that 
$$
\lambda_{j}(x,[\mathbf{w}^{(n)}_{i}]) \lambda_{j}(y,[\mathbf{w}^{(n)}_{i}])\to \mathbf{1}\{x,y\in D_{j}\}
$$
pointwise.
$\blacksquare$
\subsection{Reasoning that HHK is Nonstationary}
In order for active learning to focus on different parts of the input space, the kernel needs to be nonstationary. The HHK is in general not stationary (thus, nonstationary). This can easily be seen in case of a completely sharp partition as introduced in Section \ref{theory}. We recall that for a kernel on $\mathbb{R}^{d}$ to be stationary, it must hold that $k(x,y)=k(x+a,y+a)$ for all $x,y\in \mathbb{R}^{d}$ and all $a\in\mathbb{R}$. In case of the HHK with sharp partitions, denoted with $k$, we consider $x,y \in D_{j}$ and $a\in\mathbb{R}$ such that $x+a,y+a \in D_{j'}$, with $j\neq j'$. If we place different stationary kernels $k_{j}$ and $k_{j'}$ into the regions with $k_{j}(x',y') \neq k_{j'}(x',y')$ for all $x',y' \in\mathbb{R}^{d}$  the characterizing equality for stationarity does not hold as $$k(x,y)=k_j(x,y)\neq k_{j'}(x,y)=k_{j'}(x+a,y+a)=k(x+a,y+a).$$
Thus, $k$ is not stationary. The property $k_{j}(x',y') \neq k_{j'}(x',y')$ for all $x',y' \in\mathbb{R}^{d}$ holds, for example, when the kernel variances of two SE kernels differ.
\section{EXPERIMENTAL DETAILS}

\subsection{Prior Parameters for HHK}
\label{priorParam}
Throughout the experiments, we use the following prior parameters for the Hierarchical Hyperplane Kernel, where the local kernels $k_{j}(x,y)$ are Squared-Exponential Kernels on $\mathbb{R}^{d}$ with lengthscales $l_{i,j},i=1,\dots,d$ and variance $\sigma_{j}^{2}$:

\begin{table}[h]
	\caption{Parameters of the Prior}
	\begin{center}
		\begin{tabular}{|c|c|c|} 
			\hline
			\textbf{Variable}& \textbf{Prior} & \textbf{Parameters} \\ 
			\hline
			$l_{i,j}$ & $\mathrm{Gamma}(\alpha,\beta)$ & $\alpha=2,\beta=2$ \\ 
			\hline
			$\sigma_{j}^{2}$ & $\mathrm{Gamma}(\alpha,\beta)$ & $\alpha=2,\beta=3$ \\ 
			\hline
			$\alpha_{j}$ & $\mathrm{Gamma}(\alpha,\beta)$ & $\alpha=6,\beta=2$\\
			\hline
			$\tilde{\mathbf{w}}_{i}$ & $\mathcal{N}(\mathbf{0},\mathbf{I})$&\\
			\hline
			$\sigma^{2}$ & $\mathrm{Exp}(\lambda)$& $\lambda=10$\\
			\hline
		\end{tabular}
	\end{center}
\end{table}
All prior parameters were chosen such that many functions $f:[0,1]^{d}\to \mathbb{R}$ have sufficient support in the resulting prior in function space.

\subsection{Dataset Preparation}
\label{dataPrep}
All three datasets/tasks contained already executed queries. The input variables were transformed to the unit interval and the output was normalized. This was mainly done for the reason that the GP priors with the described prior parameters have support over the dataset, for the HHK model, but also for the TreedGP and RBF model (all models with priors on the kernel parameters). We don't see that as a restrictive assumption for real-world settings as often upper and lower bounds for input and output values are given and one might either rescale input and output variables or rescale the prior parameters. 
\subsection{Inference Time Comparision}
\begin{figure}[b]
	\centering
	\includegraphics[width=0.9\linewidth]{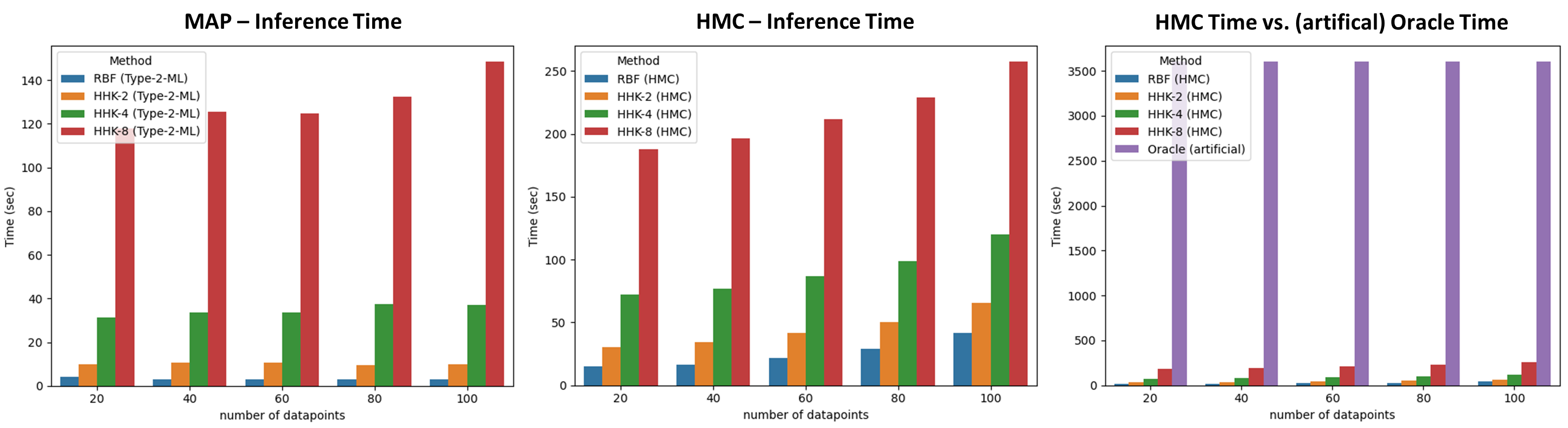}
	\caption{We show the computation time for model inference for the different HHK configurations and the RBF kernel on the LGBB dataset for different number of datapoints. In the left plot, MAP inference is shown, where the maximization is repeated ten times for inference. In the middle, the inference time for HMC is shown and in the right plot the same bars are shown with an additional bar that indicates an artificial oracle time of 1h in the active learning cycle. }
	\label{fig:inferencetimeanalysis}
\end{figure}
In Figure \ref{fig:inferencetimeanalysis}, we show the inference times for the different kernels (HHK + RBF) and for the different inference schemes (HMC and MAP) on the LGBB dataset for different number of datapoints. For MAP inference, we do an optimization with ten restarts to avoid local maxima. We note that in practice this could be parallelized in order to decrease inference times. As the number of parameters in the model increases with the number of hyperplanes, it is clear that also the inference time increases with more hyperplanes. This holds for MAP as well as for HMC as can be seen in Figure \ref{fig:inferencetimeanalysis}. However, we note that the impact of the inference time on the active learning cycle is only relevant if it makes up a significant amount of the oracle time. To illustrate that we show in the right plot of Figure \ref{fig:inferencetimeanalysis} a comparision of the inference times to an artificial oracle time of 1h. In this example the inference time would not have a big impact on the active learning performance.

\subsection{Parameterization of Warped Multi-Index Kernel}
We use the parameterization of the Warped Multi Index Gaussian process (WaMI) model described in the PhD thesis of S\'{e}bastien Marmin (\cite{marmin:tel-01743815}, p.45-47). The parameterization given in \cite{WaMIGp} can only be used for two-dimensional problems. Inside the described model in \cite{marmin:tel-01743815} they use the Beta distribution as warping function. For computational reasons we used the almost identical Kumaraswamy distribution as a replacement, which has a simpler parameterization that can be easily differentiated with automatic differentiation software (this is a commonly used alternative to the Beta distribution also used for example in the \textit{Spearmint} package for Bayesian Optimization). 

\subsection{Details on DeepGP experiments}
\label{DeepGPimplementation}
We used the R-package \textit{deepgp} associated with the method in \citet{sauer2020active} which implements DeepGPs with Slice-Sampling inference and the calculation of the ALC acquisition function. We chose a two-layer DeepGP for the experiments as it showed stable and good results in \citet{sauer2020active}. The ALC criteria needs a reference input-set $\mathcal{X}_{ref}$. All three datasets/tasks contained already executed queries (see \ref{dataPrep}) with input locations that were either uniform random (Exponential2D, ClosePI) or a grid of input locations (LGBB). We used the complete set of input locations as reference input-set $\mathcal{X}_{ref}$.

\subsection{Approximation of the Predictive Entropy}
\label{quadrature}
The estimate of the predictive distribution $p(y|x)$ given by HMC is a mixture of Gaussians:
$$p(y|x)=\frac{1}{n}\sum_{i=1}^{n}\mathcal{N}(y;\mu_{i}(x),\sigma_{i}^2(x))$$
As there is no analytical formulation for the Entropy $H(y|x)=\int p(y|x) \mathrm{log}(p(y|x)) dy$ of a mixture of Gaussians we need to approximate this quantity/integral. First, the Gaussian mixture is a density in 1D which already simplifies computations. Still, we evaluated different approximations: sampling-based approximations, first- and second Taylor approximations and quadrature. We found quadrature to have the best cost-to-precision ratio and used the SciPy implementation for 1D integrals with integration bounds $\mathrm{min}_{i=1,\dots,n}{\{\mu_{i}-2\sigma_{i}\}}$ and $\mathrm{max}_{i=1,\dots,N}{\{\mu_{i}+2\sigma_{i}\}}$ (the two-sigma quantiles of the left and right-most MCMC samples) in order to concentrate the quadrature to the regions with higher density. We also think a reason for the quadrature to work accurately in this case is that the predictive distribution has an almost Gaussian-like shape, at least if a sufficient number of MCMC samples is used. Lastly, we note that doing quadrature still can have a computational overhead if it is done in sequence over a set of evaluation points. Thus, in case of shorter oracle times, it is recommended to parallelize this computation over several CPU cores.

\label{inferencetime}
\vfill

\end{document}